\documentclass[a4paper, 10pt, conference]{ieeeconf}      
\usepackage{FG2020}
\usepackage{caption}
\usepackage{subfigure}

\FGfinalcopy 
\IEEEoverridecommandlockouts 
\usepackage{graphics} 
\usepackage{mathptmx} 
\usepackage{times} 
\usepackage{amsmath} 
\usepackage{amssymb}  

\def\FGPaperID{57} 

\usepackage[ruled, linesnumbered]{algorithm2e}
\let\labelindent\relax
\usepackage{enumitem}

\usepackage[acronym]{glossaries}
\usepackage{sidecap}
\usepackage{wrapfig}
\usepackage{tabularx, booktabs, ragged2e}
\usepackage{adjustbox}
\usepackage{multirow}
\usepackage{makecell}
\usepackage{gensymb}
\usepackage{comment}
\usepackage{xr}
\usepackage{lipsum}

\usepackage{mathrsfs}
\usepackage[pagebackref=true,breaklinks=true,letterpaper=true,colorlinks,bookmarks=false, citecolor=blue]{hyperref}

\newcommand*\rot{\rotatebox{74}}

\newcommand{\ra}[1]{\renewcommand{\arraystretch}{#1}}

\newcommand*\eg{\textit{e.g.}}
\newcommand*\ie{\textit{i.e.}}

\title{\LARGE \bf
EV-Action: Electromyography-Vision Multi-Modal Action Dataset}

\author{\parbox{16cm}{\centering
    {\large Lichen Wang$^1$, Bin Sun$^1$, Joseph Robinson$^1$, Taotao Jing$^1$, and Yun Fu$^{1,2}$}\\
    {\normalsize
    $^1$ Department of Electrical and Computer Engineering, Northeastern University, USA\\
    $^2$ Khoury College of Computer Science, Northeastern University, USA}}
    \thanks{We thank our volunteers, Allyson Vakhovskaya, Daniel J. Peluso, Emily Freed, Kasey Lee, Yue Bai, and Yunyu Liu from Northeastern University for their substantial contributions to our project in data collection, labeling, and analytical procedures.}
}

\begin{document}

\ifFGfinal
\thispagestyle{empty}
\pagestyle{empty}
\else
\author{Anonymous FG2020 submission\\ Paper ID \FGPaperID \\}
\pagestyle{plain}
\fi
\maketitle

\begin{abstract}
Multi-modal human action analysis is a critical and attractive research topic. However, the majority of the existing datasets only provide visual modalities (\ie, RGB, depth and skeleton). To make up this, we introduce a new, large-scale EV-Action dataset in this work, which  consists of RGB, depth, electromyography (EMG), and two skeleton modalities. Compared with the conventional datasets, EV-Action dataset has two major improvements: (1) we deploy a motion capturing system to obtain high quality skeleton modality, which provides more comprehensive motion information including skeleton, trajectory, acceleration with higher accuracy, sampling frequency, and more skeleton markers. (2) we introduce an EMG modality which is usually used as an effective indicator in the biomechanics area, also it has yet to be well explored in motion related research. To the best of our knowledge, this is the first action dataset with EMG modality. The details of EV-Action dataset are clarified, meanwhile, a simple yet effective framework for EMG-based action recognition is proposed. Moreover, state-of-the-art baselines are applied to evaluate the effectiveness of all the modalities. The obtained result clearly shows the validity of EMG modality in human action analysis tasks. We hope this dataset can make significant contributions to human motion analysis, computer vision, machine learning, biomechanics, and other interdisciplinary fields.
\end{abstract}

\begin{figure*}
\centering
\includegraphics[width=170mm]{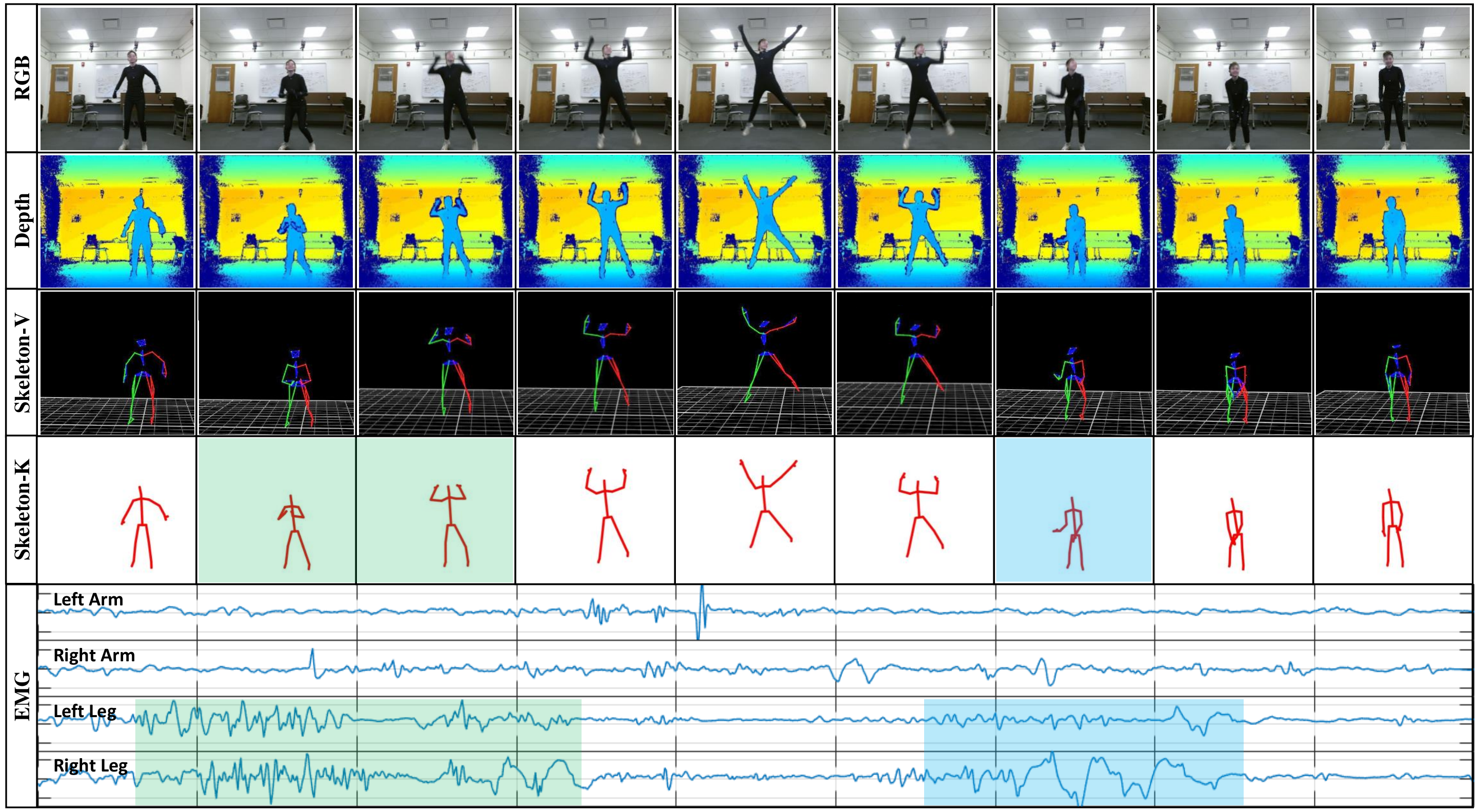}
\vspace{-1mm}
\caption{Visualization of sample frames in EV-Action dataset. Colored boxes show the correlations between visual modalities and EMG (\ie, \textit{Take Off} and \textit{Touch Down}). We can clearly observe that EMG responds early and last longer than visual modalities which provides unique view for action analysis. All modalities were well aligned and labeled.}\label{fig:samples}
\vspace{-4mm}
\end{figure*}

\begin{table*}[!tp]
\begin{center}
\caption{Technical specifications of the sensors used in EV-Action dataset.}\label{table:sensor_sum}
\vspace{-2mm}
\scalebox{0.97}{
\begin{tabular}{l c c c c c c c c}
\hline
Sensors & Modality & Resolution & Frame Rate (fps/Hz) & Skeleton Joints & Field of View & Sensor Number & Range &  Sensitivity\\
\hline
\multirow{3}{*}{Kinect-V2}  & RGB & $1,920 \times 1,080$ & $30$ & - & $84.1^\circ \times 53.8^\circ$ & 1 & - & 8-bit\\
 & Depth & $512 \times 424$ & $30$ & - & $70.6^\circ \times 60.0^\circ$ & 1 & 0.5-4.5 m & 16-bit\\
 & Skeleton-K & - & $30$ & $26$ & - & - & - & -\\
Vicon-T40s & Skeleton-V &  $2,336\times1,728$ & 100 & 39 & $98.1^\circ \times 50.1^\circ$ & 8 & 12 m & 10-bit\\
Delsys-Trigno & EMG & - & $1,000$ & - & - & 4 & $\pm$ 22 mV & 16-bit\\
\hline
\end{tabular}
}
\end{center}
\vspace{-6mm}
\end{table*}

\section{INTRODUCTION}
There are a wide range of applications for human motion analysis (\eg, event detection, behavior prediction, gait analysis, joint mechanics, prosthetic designs, sports medicines~\cite{RGB_Recog2,3dcnn,lichen_ICCV,RGBD_HUDA,lichen_seg1,lichen_seg2,lichen_MML1,lichen_AI20}). The availability of datasets tends to directly impact the progress of research. From the start, action datasets only consisted RGB modality~\cite{RGB_Recog2}. Later on, as 3D sensors became more accessible, several datasets included the depth modality~\cite{RGBD_Survey,MSRAction3D, RGBD_data2,RGBD_data4}. This paved a way for researchers to propose more effective approaches in terms of multi-modal methods~\cite{3dcnn,lichen_ICCV,RGBD_HUDA}. After that, skeleton data was introduced by some works~\cite{SK_Recog1,SK_Recog2}. However, most skeleton information of these datasets was directly obtained from Kinect sensors~\cite{kinect_v1v2}, resulting in low localization accuracy. Skeleton modal captured by more accurate devices was released~\cite{Multi_data_CMU}, while RGB-D modals were not included.

We introduce EV-Action dataset, which includes all visual modalities mentioned above (\ie, RGB, depth, and two skeleton modalities). An optical tracking-based Vicon system~\cite{vicon_VS_kinect1} is deployed to capture high-quality skeleton motion information. Compared with Kinect, Vicon achieves significantly higher sampling rate ($100$ vs. $30$ fps), higher localization accuracy, and more skeleton markers ($39$ vs. $26$). It provides more comprehensive skeleton motion information in terms of location, trajectory, velocity, and acceleration. We further collected Electromyography (EMG) signals to measure the electrical activity of human skeletal muscles as a function of the intensity of force~\cite{EMG_intro}. EMG is regularly used in medical and biomechanics fields. It has not yet been well explored in the fields of human motion analysis. In EV-Action, all modalities are captured simultaneously with action labels frame by frame. The goal of EV-Action is exploring the latent correlation across different modalities and improving the performances of action analytic tasks. EV-Action could contribute significantly to the research fields of human motion analysis, multimedia, computer vision, machine learning, biomechanics, and other interdisciplinary sub-fields. The contributions of our paper are shown below:
\begin{enumerate}
\item We designed and constructed a data collection center with optical tracking system and Kinect-V2 systems. This allowed us to capture the four visual modalities (\ie, RGB, Depth, Skeleton-K, and Skeleton-V).
\item EMG signal from skeletal muscles is extracted. This is the first action dataset including EMG, which provides complimentary information and reveals valuable correlations between visual and non-visual modalities.
\item A simple yet effective EMG recognition framework is proposed which achieves highest performance and reveals unique characteristics of EMG in human actions.
\item We defined experimental settings and provided the state-of-the-art benchmarks for each modality. EMG is merged with other modalities which further demonstrates the complementary of the EMG modal.
\end{enumerate}

\section{RELATED WORKS}

\subsection{RGB/D and Skeleton Datasets}
Small-scale datasets included tens of action classes (\eg, Weizmann~\cite{weiz_dataset}) are initially deployed for  action analytic tasks~\cite{RGBD_Survey}. Upon the arrival of deep learning, large-scale RGB datasets were introduced (\eg, UCF101~\cite{RGB_UCF101} and Kinetics~\cite{RGB_kinetics}). Later on, RGB-D datasets were released (\eg, MSR-Action3D~\cite{MSRAction3D}, RGBD-HuDaAct~\cite{RGBD_data2}). Due to the space and budgeting constraints, most RGB-D datasets were collected using low-cost Kinect sensors~\cite{kinect_v1v2,kinect_book}. In addition, Kinect sensors can extract skeleton data, as introduced in MAD~\cite{MAD1}, UCF-Kinect~\cite{SK_data_UCF} and NTU-RGBD~\cite{SK_data_UTN}. However, the accuracy and stability of Kinect are low, which limits the potential research of action analysis.

\subsection{Multi-Modal Action Datasets}
We consider the dataset containing more than RGB-D modalities as multi-modal dataset. Currently, only a few datasets provide additional modalities. NTU-RGBD~\cite{SK_data_UTN} and PKU-MMD~\cite{PKU_MMD} contained infrared frames captured by Kinect sensors. CMU-MMAC~\cite{Multi_data_CMU} utilized an optical tracking technique to capture action sequences. UTD-MHAD~\cite{Multi_data_UT} utilized a single wearable inertial sensor to capture inertial signals. However, the modality is severely limited and sporadic due to the inconsistent collection manner. Our EV-Action dataset utilizes $39$ markers to capture precise location, trajectory and acceleration information at a high frame rate (100 fps). To this end, EV-Action is the \textbf{most accurate and comprehensive} dataset of this kind.

\subsection{EMG Signal}
Electromyography (EMG) is an electrodiagnostic technique to evaluate the electrical activity produced by skeletal muscles~\cite{EMG_ReBio1,EMG_intro}. Typically, EMG is used in neural science, biomechanics, and signal processing fields (\eg, hand gesture~\cite{EMG_handGesture}, robot arm control~\cite{EMG_Arm}, face expression~\cite{EMG_face1}. Since EMG activates before visual signal which could foresee potential information such as intention, force, and even mental activities information that cannot be recognized in visual domain. To this end, we consider EMG as another crucial clue for exploring actions. To the best of our knowledge, no existing work associates EMG with other action modalities. Considering the potential applications that could be had, we generated EV-Action with EMG as one of the critical modalities for human action analysis.

\begin{figure*}[!tp]
\centering
\includegraphics[width=177mm]{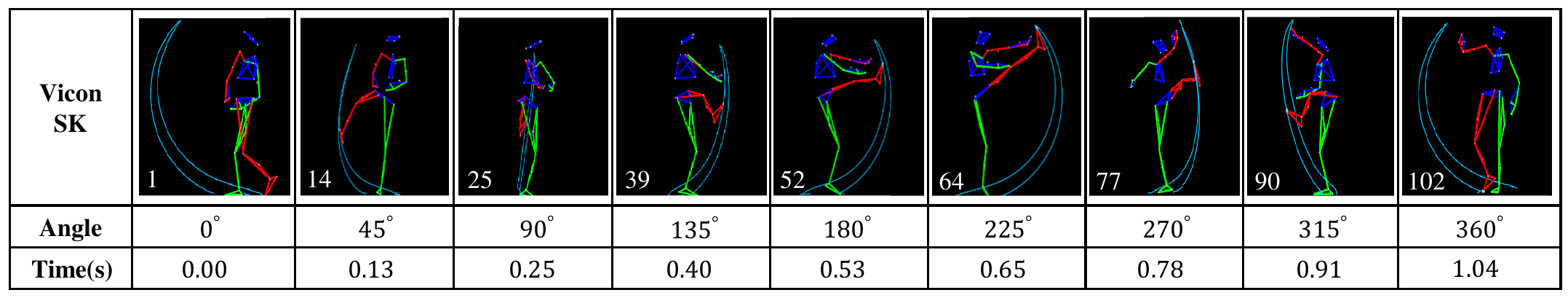}
\vspace{-6mm}
\caption{Visualization of a subject performing a kicking action across view angles and time. The blue curve highlights the trajectory of a marker. Clearly, EV-Action contains the precise detailed motion information of the actions. Frame numbers are shown in left bottom which indicate the high sampling rate of Vicon system.}\label{fig:angles}
\vspace{-4mm}
\end{figure*}

\begin{figure}[!tp]
\centering
\subfigure[]
{
   \includegraphics[width=74mm]{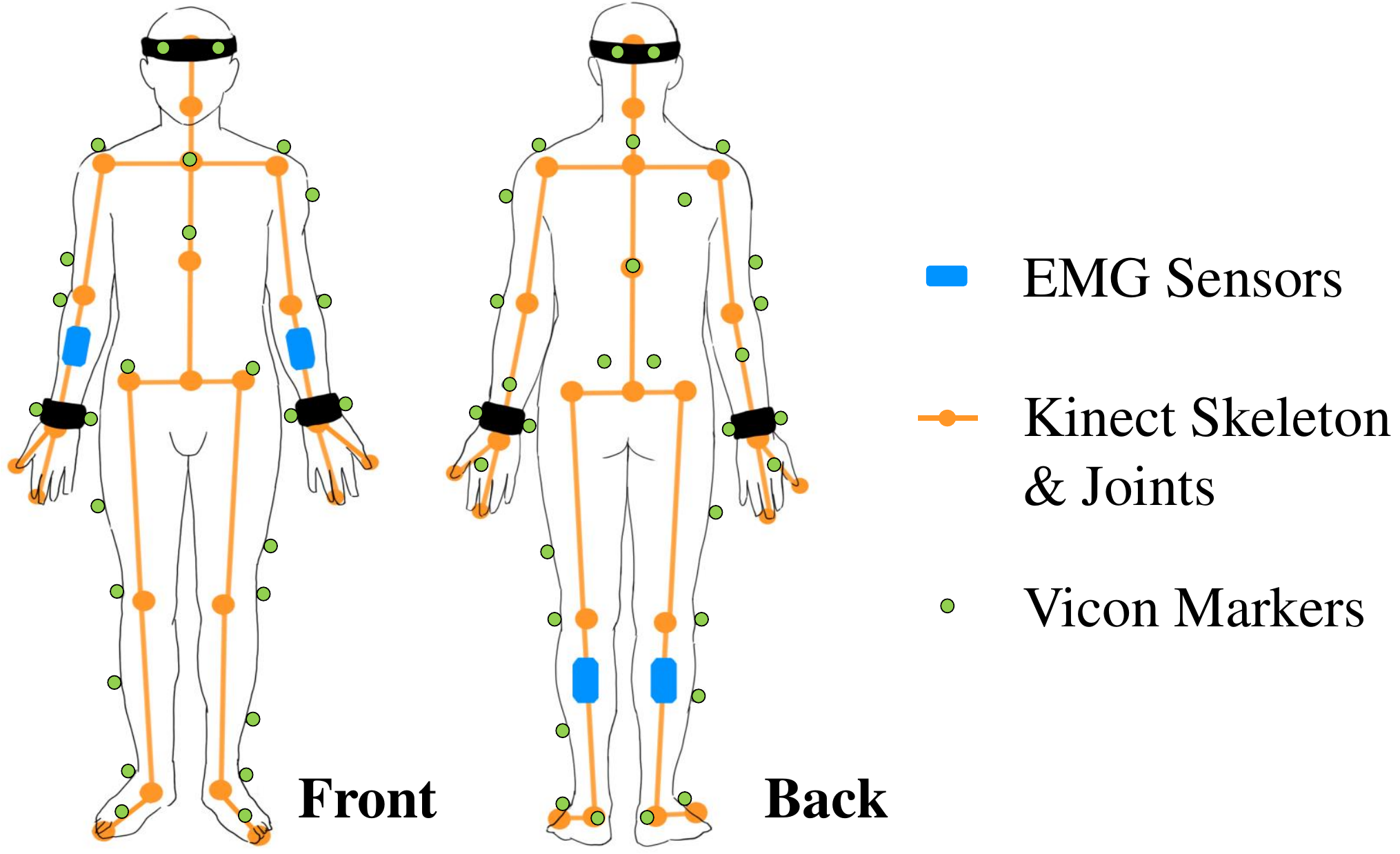}\label{fig:vicon_place}
}
\vspace{-2mm}
\subfigure[]
{
   \includegraphics[width=75mm]{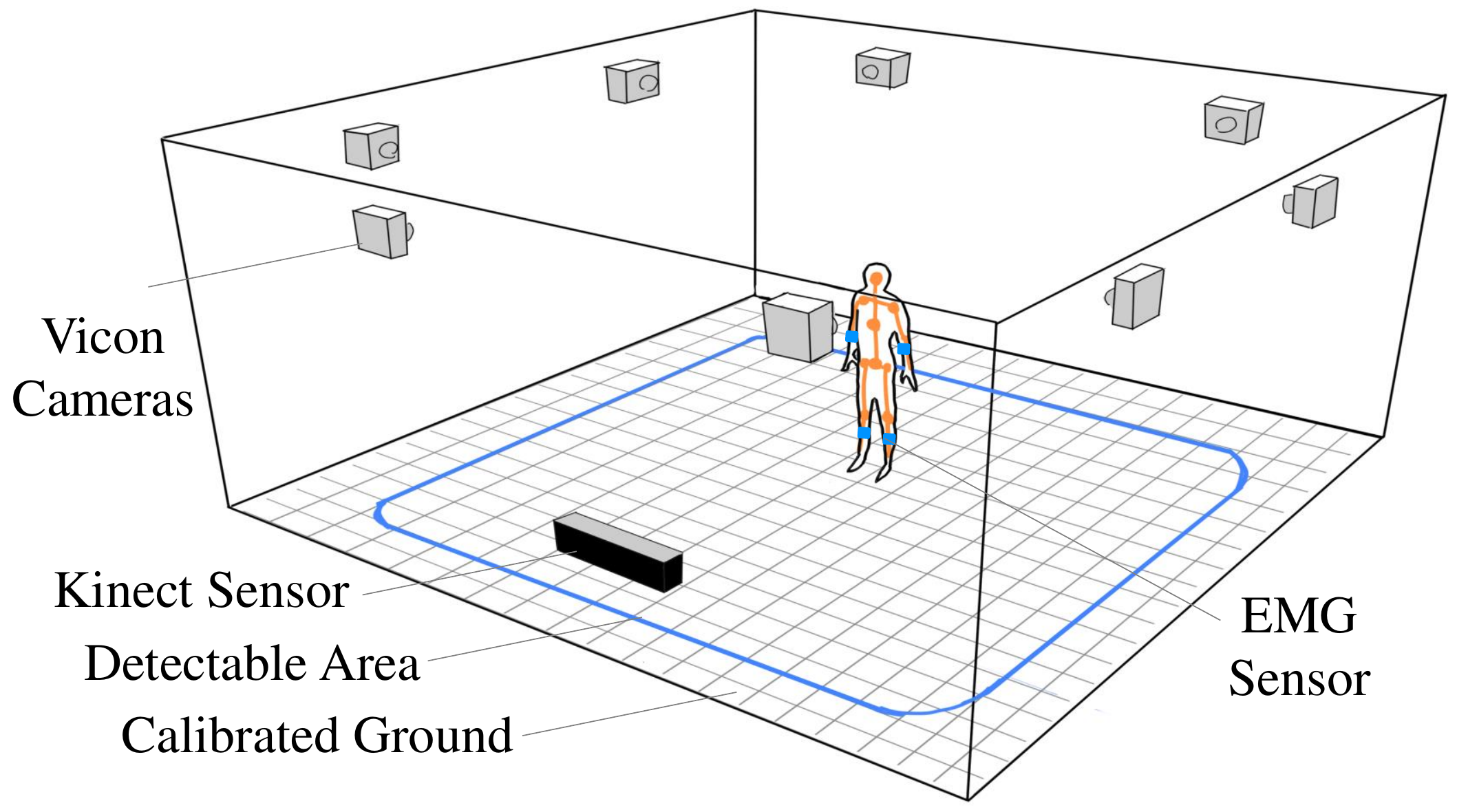}\label{fig:env_set}
}
\vspace{-1mm}
\caption{(a) Sensor placement schemes. Orange lines and spots indicate Kinect skeleton with 26 joints. Small gray points denote Vicon markers. And blue blocks indicate the EMG sensors. (b) Data collection center environment setup.}\label{fig:3places}
\vspace{-4mm}
\end{figure}

\section{EV-ACTION DATASET}

\subsection{Sensors and Setup}
There are 1 Kinect-V2 sensor~\cite{kinect2_acc}, 4 wireless EMG sensors, and 8 Vicon-T40s cameras in the data collection system.

\textbf{Kinect}~\cite{kinect2_acc} captures RGB-D modalities from subjects. Skeleton information is further extracted from the depth image. We used a second generation Kinect~\cite{kinect2_acc,kinect_v1v2} (Kinect-V2) which has a high resolution camera (1,920$\times$1,080) at 30 fps with a wide field of view (70$^\circ \times$60$^\circ$). Moreover, the resolution of the depth sensor is 512$\times$424. It is more robust and efficient for pose estimation with reference to 26 joints (Figure~\ref{fig:vicon_place}). In the collection procedure, a Kinect-V2 captures the subjects in the front view (Figure~\ref{fig:env_set}).

\textbf{Vicon System} utilizes optical tracking-based technology to capture skeleton data with more accurate and comprehensive motion information~\cite{vicon_VS_kinect1}. We deploy 8 Vicon-T40s infrared cameras to capture the stickup marks on each subject (Figure \ref{fig:vicon_place}). The cameras sample data points as 10-bit gray-scale frames at 100 fps and with a resolution of 2336$\times$1728. Then, the frames were calibrated and labeled to obtain skeleton information. We follow the standard scheme~\cite{vicon_plugin} by placing 39 markers around human body (Figure~\ref{fig:vicon_place}). It captures precise and comprehensive motion information, such as the second bounce in the \textit{Fall Down} action class. Also, due to the high frame rate and accuracy, high quality trajectories and accelerations were obtainable in reference to ground coordinates. Figure~\ref{fig:angles} shows the \textit{Kick} action viewed across time and at different angles, with the blue curve indicating the trajectory of the toe marker. No other action datasets provides such detailed information.

\textbf{EMG Sensor} captures EMG signals from human muscles. We deploy wireless EMG sensors which captures 16-bit EMG signal at 1000 Hz. This enables the sensors to cover the whole frequency spectrum of skeletal EMG (\ie, 20-450 Hz) signal. We attached 4 sensors to each subject: the middle of each forearm and the shank muscles (Figure \ref{fig:vicon_place}). There are 3 reasons: 1), the common actions usually utilize arms and legs; 2), the location of each muscle (mid-line of the muscle in the belly that is between the myotendinous junction and the nearest innervation zone) gives off a signal of highest amplitude, which makes the signal most responsive to the corresponding action~\cite{EMG_intro}; 3), the \textit{crosstalk} noise generated by neighboring muscles has the potential to get misinterpreted for originating from a muscle of interest, and placing the sensor mid-line makes it less susceptible to this noise.

\begin{figure}[t!]
\centering
\subfigure[]
{
\includegraphics[width=40mm,height=27mm]{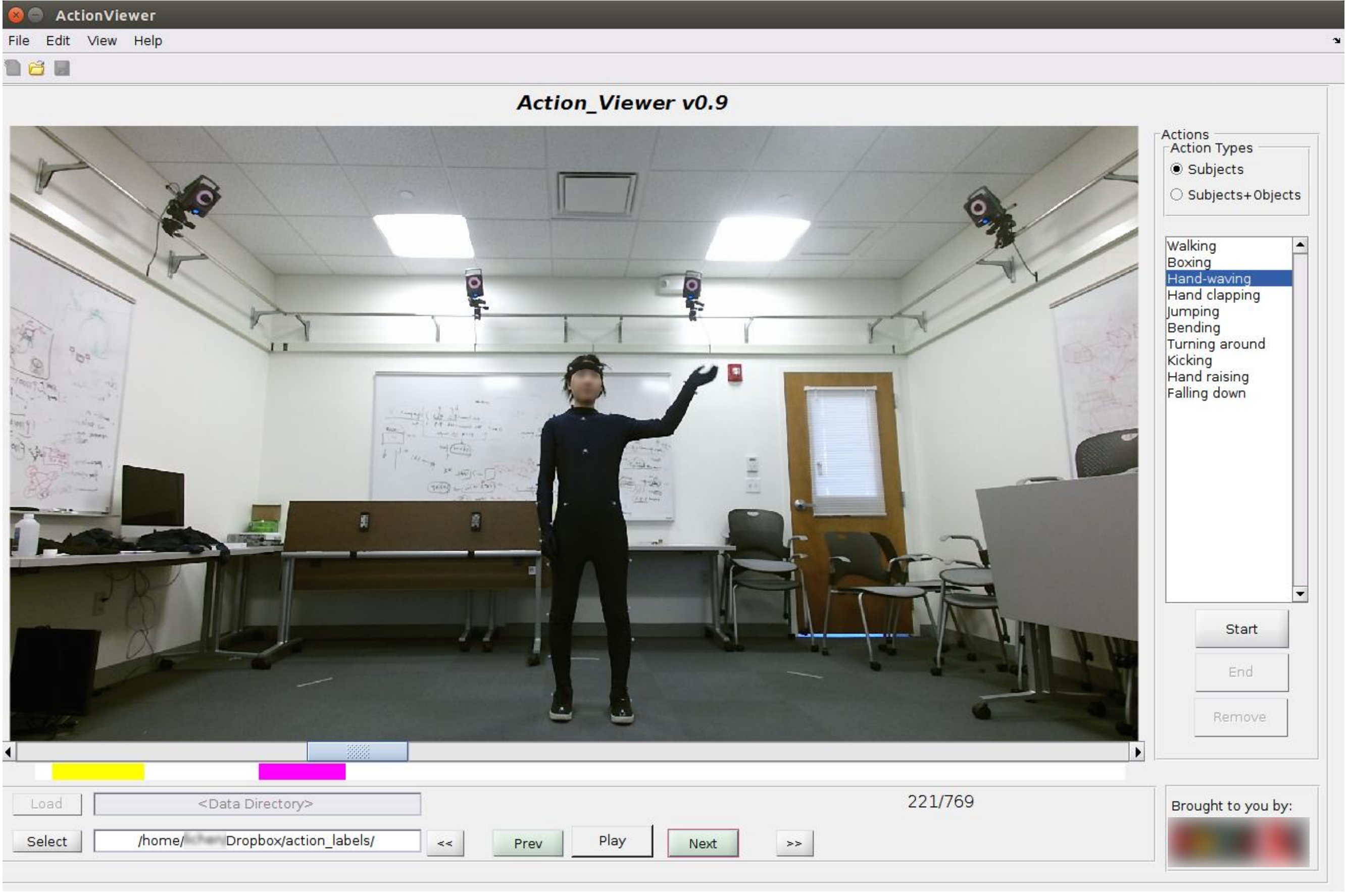}\label{fig:gui}
}
\subfigure[]
{
\includegraphics[width=40mm,height=27mm]{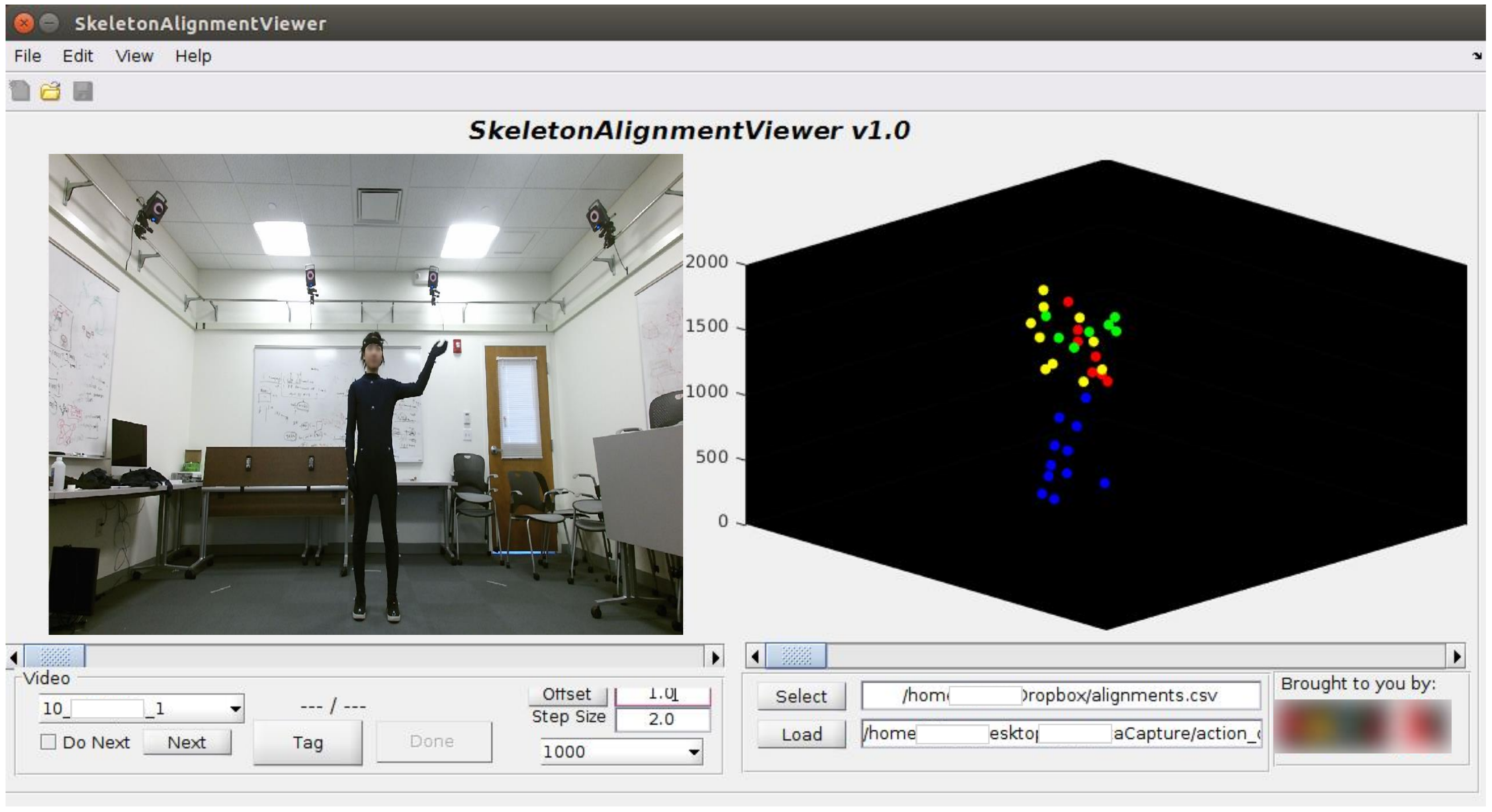}\label{fig:alignmentGUI}
}
\vspace{-4mm}
\caption{(a) Video labeling tool used to label action sequences, and (b) tool used to precisely align modalities.}
\vspace{-5mm}
\label{fig:subfig} 
\end{figure}

\textbf{Data Collection Center} consists of 8 the Vicon cameras placed around the parameter of a $\textup{4.6m} \times \textup{4.6m}$ room which has a detectable area of $\textup{3m} \times \textup{3m}$. All traceable markers fell in the Vicon cameras field of view. There was a single Kinect sensor centered facing front, and each action was performed with the face of this Kinect sensor as the front. 4 EMG sensors were connected to each subject (Figure \ref{fig:env_set}).

\setlength{\tabcolsep}{4pt}
\begin{table*}[!tp]
\begin{center}
\caption{Comparison between EV-Action dataset and other popular multi-modal datasets. EV-Action is one of the largest multi-modal datasets and significantly outperforms other datasets in modal diversity, subject numbers, and sample clips.}\label{table:actionNum}
\vspace{-2mm}
\scalebox{0.95}{
\begin{tabular}{l c c c c c c}
\hline
Datasets & Samples & Classes & Subjects & Framerate (fps)  & Sensors & Modalities\\
\hline
RGBD-HUDA~\cite{RGBD_data2} & 1189 & 13 & 30 & 30 & KinectV1 & RGB+D\\
MSR-Action3D~\cite{MSRAction3D} & 567 & 20 & 10 & 30  & RGB-Cam & D+SK\\
CAD-60~\cite{RGBD_HUDA} & 60 & 12 & 4  & 30 & KinectV1 & RGB+D+SK\\
Action4$^2$~\cite{action42} & 6844 & 14 & 24  & 30 & KinectV1 & RGB+D\\
CAD-120~\cite{RGBD_data4} & 120 & 20 & 4  & 30 & KinectV1 & RGB+D+SK\\
Multiview 3D Event~\cite{3DEvent} & 3815 & 8 & 8  & 30 & KinectV1 & RGB+D+SK\\
Online RGB+D Action~\cite{onlineRGB} & 336 & 7 & 24  & 30 & KinectV1 & RGB+D+SK\\
Northwestern-UCLA~\cite{northeastern_UCLA} & 1475 & 10 & 10  & 30 & KinectV1 & RGB+D+SK\\
UWA3D Multiview~\cite{UWA_3D} & 900 & 30 & 10  & 30 & KinectV1 & RGB+D+SK\\
Office Activity~\cite{office_3d} & 1180 & 20 & 10  & 30 & KinectV1 & RGB+D\\
UTD-MHAD~\cite{Multi_data_UT} & 861 & 27 & 8  & 30+50 & KinectV1+WIS & RGB+D+SK\\
3D Action Pairs~\cite{3DActionPair} & 360 & 12 & 10  & 30 & KinectV1 & RGB+D+SK\\
UWA3D Multiview II~\cite{UWA_3D2} & 1075 & \textbf{30} & 10 & 30  & KinectV1 & RGB+D+SK\\
EV-Action (Ours) & \textbf{7000} & 20 & \textbf{70}  & \textbf{30+100+1000} & \textbf{KinectV2+Vicon+EMG} & \textbf{RGB+D+SKK+SKV+EMG}\\
\hline
\end{tabular}
}
\end{center}
\vspace{-4mm}
\end{table*}

\subsection{Dataset Description}
Completeness, comprehensiveness, and diversity were highly considered when building EV-Action. To make it practical and generalizable, we included 20 common actions (Table~\ref{table:actions}), 10 were done by a single subject and the other 10 were done by that same subject interacting with different objects. The dataset includes 70 subjects performing the actions 5 times (\ie, 100 action clips per subject). To introduce diversities, the subjects intentionally perform slightly different style in each loop. All-in-all, resulting in $7000$ action clips at multiple views. Table~\ref{table:actionNum} summarizes these statistics compared with recent and popular multi-modal action datasets. It is clear that EV-Action is one of the largest multi-modal datasets, as it significantly surpasses other datasets in terms of modal diversity, number of subjects, and number of samples. And it includes non-visual EMG signal for the first time. Referencing figure~\ref{fig:samples}, we notice that the EMG signal activates prior to the actions (\ie, \textit{Take Off} and \textit{Touch Down}). We also notice that the duration of EMG are typically longer than the visual modalities. These patterns are unrecognizable from any visual modal. It demonstrates that EMG does provide unique and complementary information for more deep and sophisticated action analytical research.

\subsection{Data Labeling}
There are two steps for data labeling: 1) annotating the actions in RGB modality, and 2) aligning the Skeleton-K and Skeleton-V. Since other modalities are captured synchronously with either Kinect or Vicon, thus, the rest of the modalities are also well aligned automatically. \textbf{Video Anotation:} We built a MATLAB labeling tool to facilitate the labeling process (Figure \ref{fig:gui}). The tool displays a video sample for a human labeler to tag start and end frames of actions selected from the predefined lists shown in Table~\ref{table:actions}. \textbf{Data Alignment:} We then align the clip with the clip captured by the Vicon system. We develop another MATLAB tool that allows us to visually align single frame of the RGB and the skeleton from the Kinect and the Vicon systems, respectively (Figure \ref{fig:alignmentGUI}).

\begin{table}[!tp]
\begin{center}
\caption{A list of the 20 actions included in EV-Action.}\label{table:actions}
\vspace{-2mm}
\scalebox{0.95}{
\begin{tabular}{l l l l}
\hline
\multicolumn{2}{c}{Single Person Actions} & \multicolumn{2}{c}{Person-Objects Actions} \\
\hline
1. Walk & 6. Bend Over & 1. Answer Phone & 6. Throw Ball\\
2. Boxing & 7. Turn Around & 2. Check Watch & 7. Drink Water\\
3. Wave Hands & 8. Kick & 3. Stand Up & 8. Tie Shoes\\
4. Clap Hands & 9. Raise Hand & 4. Sit Down & 9. Read Book\\
5. Jump & 10. Fall Down & 5. Grab Bag & 10. Move Table\\
\hline
\end{tabular}
}
\end{center}
\vspace{-7mm}
\end{table}

\section{DATA ANALYSIS}
\begin{figure}[!tp]
\centering
\includegraphics[width=80mm]{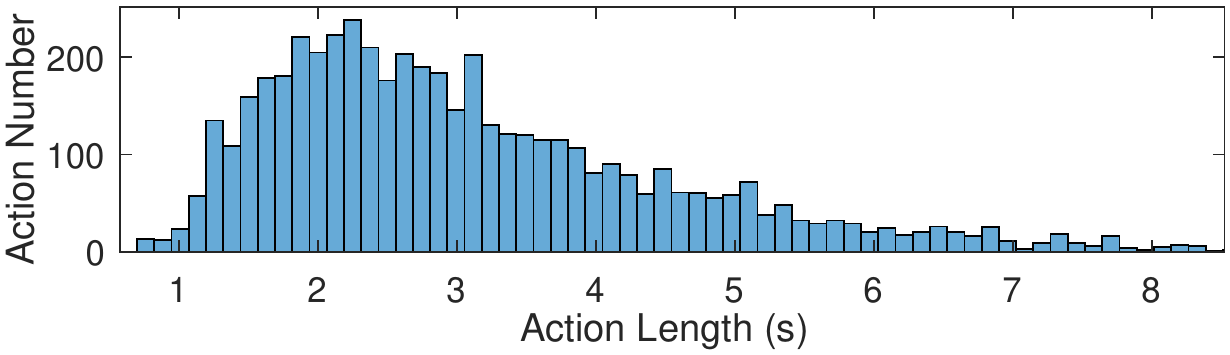}
\vspace{-2mm}
\caption{Histogram of the length distribution of all videos.}\label{fig:lengthHist}
\vspace{-4mm}
\end{figure}

\begin{figure}[!tp]
\centering
\includegraphics[width=80mm]{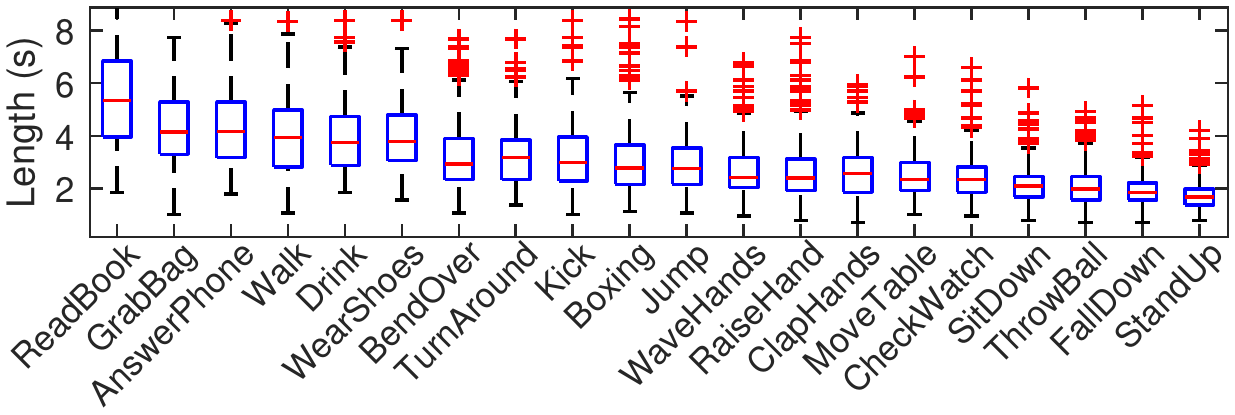}
\vspace{-1mm}
\caption{Video length distribution of each action.}\label{fig:lengthBox}
\vspace{-4mm}
\end{figure}

Histograms of all video length are shown in Figure~\ref{fig:lengthHist}, and box plots (Figure \ref{fig:lengthBox}) depict action-specific statistics from longest to shortest. We observe that the video lengths for different actions varied. For instance, \textit{Read Book} tends to be the longest, with \textit{Stand Up} the shortest. Moreover, variation in video length exists for the same action across different subjects, which is especially true for repetitive actions. For instance, when preforming actions such as \textit{Boxing} or \textit{Jump}, subjects prefer to choose the exact number of reps. For non-repetitive actions there tended to be less variation across different subjects. Since subjects perform different actions continuously without intentional pause during collection; another observation is that subjects tended to move faster between actions. This sometimes results in the end of the current action and the beginning of the next getting mixed across frames in between (\ie, overlap between actions). For example, when a subject \textit{Put Down Phone} and then \textit{Check Watch} immediately, there might be an overlap. These situations make more of challenges while make it better suited for more practical research.

\begin{figure}[!tp]
\centering
\includegraphics[width=81mm]{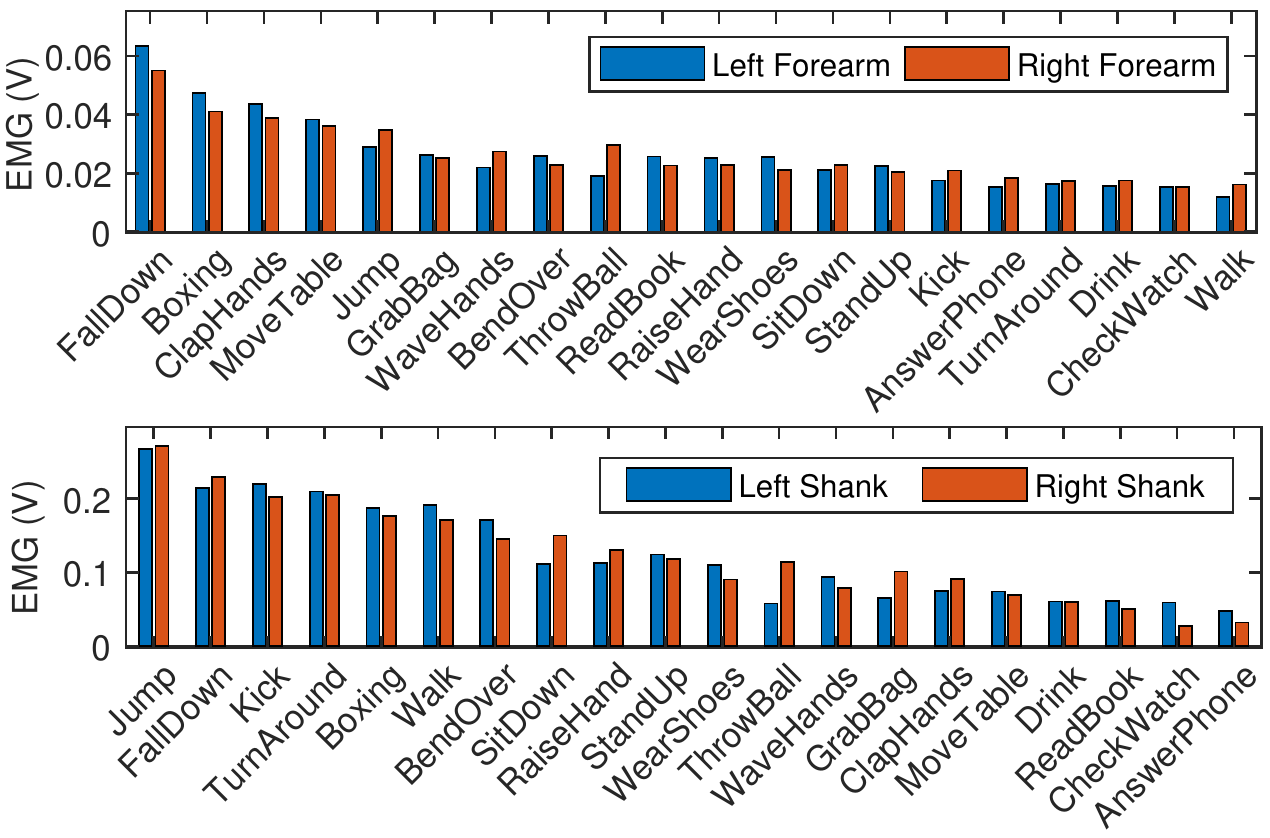}
\vspace{-2mm}
\caption{The average of Root Mean Square (RMS) value of the EMG recordings in different actions. We separate the value of upper body (left and right forearm) and lower body (left and right shank) for better discussion.}\label{fig:emg_sort}
\vspace{-3mm}
\end{figure}

Root Mean Squared (RMS) is an effective method to pre-process EMG data~\cite{EMG_intro}. We obtain the average of action RMS and surprisingly notice that the shank muscles have significantly higher (2 times) amplitude than forearm muscles since the stronger and bigger muscles around the shank region. We separately illustrated these four channels in Figure~\ref{fig:emg_sort} and found more interesting observations. For instances, most subjects utilize right hand for \textit{Throwing Ball}, while they are also utilizing their right legs simultaneously (might for balance requirement). Moreover, subjects use left legs even for \textit{Check Watch} (might for body balance; in order to rise left arm to check watch, they should hold/balance their body by left leg to take over left arm). These observations are unique and valuable for action understanding but cannot be obtained in any visual modality. We wish more interesting discoveries could be revealed by exploring EMG modal.

Since Vicon tracks the markers pasted on subjects, there were situations that several markers were obstructed to the cameras, such as \textit{Fall Down} and \textit{Sit Down}. Once the occluded marker is again detected, Vicon could re-localized the respective point. In response for the missing situation, we split the data as two types, unlabeled marker locations data and labeled skeleton data. Thus, advanced skeleton reconstruction methods or label independent research can also explore EV-Action. This also leads to believe that more sophisticated algorithms are needed to achieve higher performance rating (\eg, missing-modality and multi-view algorithms). The rest modalities (\ie, RGB, Depth, Skeleton-K and EMG) are stable across all actions without noticeable errors.

\section{EXPERIMENTS}
State-of-the-art approaches were used to benchmark the different modalities. Specifically, single-modal benchmarks using RGB, Skeleton-K, Skeleton-V were done. In the multi-modal scenario, RGB-D, Skeleton-K + EMG, and Skeleton-V + EMG were conducted. We achieved considerable performance improvements by employing a simple, yet effective fusion technique (\ie, fused at the feature-level). This project is the first to model the non-visual EMG signal for action recognition. This is also a great promise for further improvement by providing more sophisticated learning frameworks and fusion techniques. Considering the information captured in an EMG signal, it is capable of discriminating between action types in itself, thus, it is complimentary to visual evidence. Thus, the EMG modality could both improve our current action recognition capabilities and serve as a necessity for certain applications.

\subsection{Experimental Settings}
Benchmarks on EV-Action followed conventional classification settings. The action clips from $56$ subjects were used during training (\ie, $5600$ clips), while the other $14$ subjects were set aside for testing (\ie, $1400$ clips). All experiments were evaluated in terms of classification accuracy.

\subsection{EMG Signal}
Signal processing methods associated with hand-crafted features are usually deployed for EMG analysis. We design a novel deep-structure framework for EMG recognition.

\begin{figure}[!tp]
\centering
\includegraphics[width=85mm]{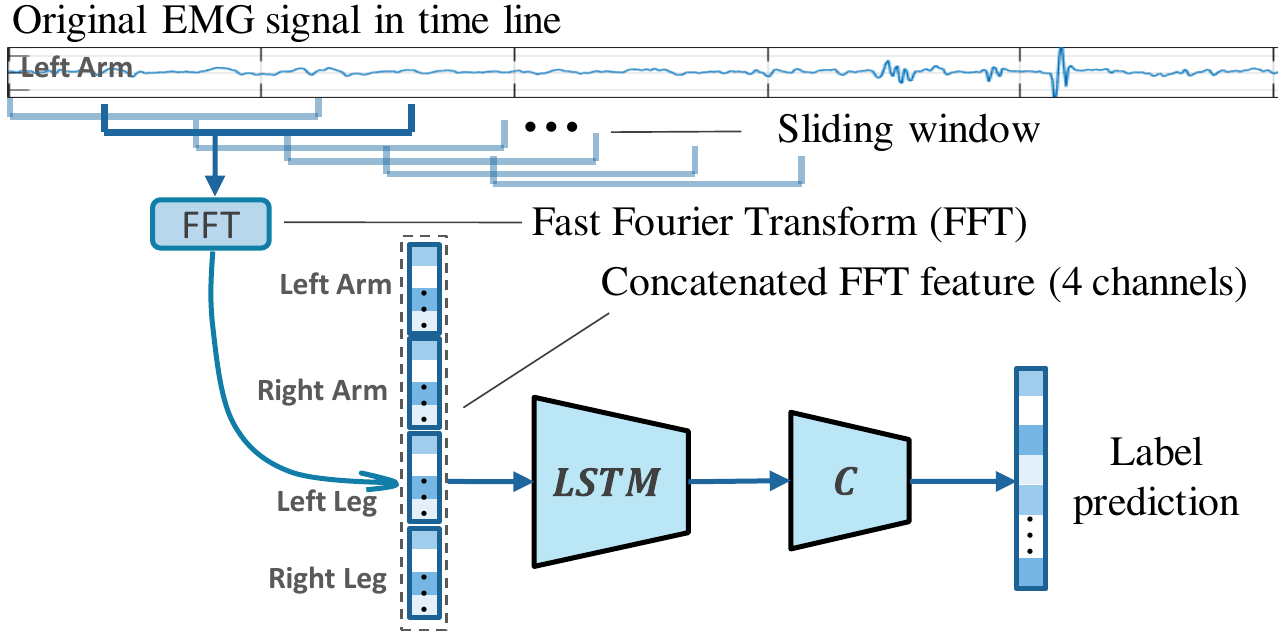}
\vspace{-1mm}
\caption{EMG modal based action recognition framework.}\label{fig:emg_framework}
\vspace{-2mm}
\end{figure}

We first introduce the conventional procedures for EMG classification. As noise of raw EMG signals occurred during collecting period, choosing the best way to extract features and reduce the dimension are crucial to achieve high classification performance. Butterworth filter~\cite{Butterworth} yields a flat frequency response which is effective to filter out EMG noise. The generalized equation is: $H_{j\omega} = (\sqrt{1+\varepsilon ^{2}(\frac{\omega }{\omega _{p}})^{2n}})^{-1}$, where $n$ is the filter order and we received the best classification performance when $n=5$. $\omega$ is the radian frequency and $\omega=2\pi f$. ${\omega _{p}}$ is the pass band frequency and $\varepsilon$ is the maximum pass band. $H$ is the frequency response. The low frequency cutoff value is set to $10$ Hz, which removes the static electricity variation caused by friction and movement. We set high frequency cutoff value at $500$ Hz, since it is the highest frequency of the EMG signal. RMS is further deployed. The expression of RMS is $R_k= (\frac{1}{N} \sum_{i=1}^{N} x_{i}^2)^{\frac{1}{2}}$, where $R_k$ is the RMS value and $x_i$ is the EMG signal of the $i$-th frame in the $k$-th time window period. $N$ is the size of sliding window. We obtain the RMS values from each channel to obtain RMS features. Linear Discriminative Analysis (LDA)~\cite{LDA} and Principal Component Analysis (PCA)~\cite{PCA} are utilized for dimension reduction. Three classifiers are tested including SVM~\cite{SVM}, K-Nearest Neighbors (KNN)~\cite{KNN}, and Random Forests (RF)~\cite{RandomForest}. The results (Table~\ref{table:EMG}) indicates that RF after PCA has the best classification performance which is $35.12\%$.

We then introduce our deep modal-based EMG action recognition approach (Figure~\ref{fig:emg_framework}). It is an effective and efficient framework in an end-to-end scenario, while both the noise elimination and the recognition are done simultaneously. Sliding windows are first employed to extract EMG signal from each channel. Differently, we utilize Fast Fourier Transform (FFT)~\cite{FFT} instead of RMS as the initial approach. This strategy has two advantages. Firstly, FFT decomposes the time series EMG data in frequency domain which automatically separates EMG with high/low noise. Thus, denoising procedures can be omitted. Secondly, FFT preserves more comprehensive information. Then, we utilize the amplitude of each frequency as feature vector, and concatenate the four channels (\ie, left/right forearm/shank) together and input them into a Long Short-Term Memory (LSTM)~\cite{LSTM} networks. LSTM outputs the feature representations. A classifier, $C(\cdot)$, is utilized to obtain the final label. The loss function is shown as below.
\begin{equation}\label{LSTM_LOSS}
  \begin{array}{cl}
    L = \|Y-C(G(F(X))\|_\textup{F}^2,
  \end{array}
\end{equation}
where $Y$ is the instance label, $X$ is the original data in time domain. $F(\cdot)$ is the FFT feature extractor. $G(\cdot)$ is an LSTM network. In the implementation, we deploy half-overlapping sliding window associated with the window size $200$ to extract data blocks. By deploying FFT to the extract block, we obtain a 100-dimension feature vector in frequency domain. Since there are four EMG channels, each data block is represented by a 400-dimension vector. The LSTM structure has a hidden layer of 1024-dimension. The result (Table~\ref{table:EMG}) denotes that the our approach significantly outperforms conventional methods which also indicates the effectiveness of EMG in action analytical tasks.

\begin{table}[!tp]
\begin{center}
\caption{EMG classification accuracy based on different dimension reduction (Dim-Red) approaches and classifiers.}\label{table:EMG}
\vspace{-1mm}
\scalebox{0.98}{
\begin{tabular}{ l  c c c }
\hline
\multirow{2}{*}{Methods} & \multicolumn{3}{c}{Dimension Reduction}\\
 & (None) & LDA & PCA\\
\hline
Random Forest & 33.72 & 16.81 & 35.12 \\
KNN & 22.16 & 13.55 & 26.18 \\
SVM & 23.74 & 16.12 & 25.65 \\
FFT-LSTM (Ours) & \textbf{44.13} & - & - \\
\hline
\end{tabular}
}
\end{center}
\vspace{-5mm}
\end{table}

\subsection{RGB \& Depth}
We evaluated single-view action recognition baselines on EV-Action. Details are introduced below:

\noindent\textbf{Action Vector of Local Aggregated Descriptor (Action-VLAD)}~\cite{ActionVLAD} is an effective video descriptor. A trainable aggregation framework is designed to capture spatio-temporal features. We fine-tune the last layer before SoftMax pre-trained by ImageNet~\cite{imagenet} our evaluation. \textbf{Temporal Segment Networks (TSN)}~\cite{TSN} sparsely samples the videos to capture the temporal information in supervised scenario. In this way, the entire video was learned effectively. \textbf{Long-term Recurrent Convolutional Networks (LRCN)}~\cite{LRCN} deploys a hierarchical visual representation learning associated with a temporal dynamic recognition module. LRCN is capable of end-to-end training. \textbf{Weighted Depth Motion Maps (WDMM)}~\cite{Dynamic3D} recognizes actions from depth videos. It utilizes a video summarization step for hierarchical representation learning. WDMM effectively increases inter-class dissimilarities and intra-class similarities. 110 is set as the number of PCA components and 80 is set as the visual words for extracting depth feature. \textbf{Weighted Hierarchical Depth Motion Maps (WHDMM)}~\cite{RGBD_wang2016action} utilizes a convolutional neural network to extract three-channel features. A hierarchical depth motion extractor is further deployed for action recognition. We only use the front view to train and test. The remaining parameters followed the original work.

\setlength{\tabcolsep}{.9pt}
\begin{table*}\centering
	\caption {Action recognition accuracy scores (\%) for all benchmarks.}
	\ra{0.8}
	\begin{tabular}{@{}cr@{\hskip .2in}cccccccccccc@{\hskip .2in}cccccccccc@{\hskip 0.08in}c@{\hskip 0.08in}ccr@{}}\toprule
		 \multicolumn{1}{c}{} & \multicolumn{10}{c}{{\ \ \ \ \ \ \ \ \ \ \ \ \ \ \ \ \ \ \ \ \ \ \ \ \ \ \ \ \ \ \ \ \ \ \ \ \ \ \ \ \ \ Single-Person}} & \phantom{a}& \multicolumn{10}{c}{{\ \ \ \ \ \ \ \ \ \ \ \ \ \ \ \ \ \ \ \ \ \ \ \ \ \ \ \ Person-Object}} &
		\phantom{c} & \multicolumn{1}{c}{}\\
		\cmidrule{3-13} \cmidrule{15-25}
		& &&\rot{Walk \ \ \ \ \ \ \ \ \ \ } & \rot{Box} &\rot{Wave Hand} &\rot{Clap Hands} &\rot{Jump}&\rot{Bend} &\rot{Turn Around}&\rot{Kick}&\rot{Raise Hand}&\rot{Fall Down}
		&&\rot{Ans. Phone} & \rot{Check Watch} &\rot{Stand Up}&\rot{Sit Down}&\rot{Grab Bag}& \rot{Throw Ball}& \rot{Drink Water}& \rot{Tie Shoes}& \rot{Read Book}&\rot{Move Table}&& ACC\\
\midrule
\multirow{3}{*}{RGB} & TSN~\cite{TSN} && \textbf{56.1} & \textbf{94.1} & \textbf{25.3} & \textbf{83.9} & \textbf{88.5} & \textbf{94.3} & \textbf{68.3} & \textbf{95.6} & 95.1 & \textbf{86.2} && \textbf{69.5} & \textbf{37.6} & \textbf{87.0} & \textbf{54.3} & \textbf{86.9} & \textbf{75.7} & \textbf{56.8} & \textbf{84.8} & \textbf{96.7} & \textbf{59.1} && \textbf{74.7}\\
& LRCN~\cite{LRCN} && 44.2 & 84.0 & 19.8 & 69.4 & 71.6 & 78.0 & 57.9 & 82.1 & 90.0 & 71.3 && 55.6 & 28.5 & 72.1 & 43.4 & 72.0 & 62.5 & 46.8 & 70.2 & 85.4 & 44.2 && 62.4\\
& VLAD~\cite{ActionVLAD} && 47.5 & 91.8 & 21.6 & 75.9 & 78.3 & 85.3 & 63.3 & 89.7 & \textbf{98.4} & 77.9 && 60.7 & 31.1 & 78.8 & 47.5 & 78.7 & 68.3 & 50.8 & 76.7& 93.4 & 48.3 && 68.2\\
\midrule 
\multirow{2}{*}{Dep} & WDMM~\cite{Dynamic3D} && 44.3 & 76.3 & 11.4 & 31.4 & 36.5 & \textbf{43.7} & \textbf{17.2} & 47.4 & \textbf{72.7} & \textbf{36.2} && \textbf{27.9} & \textbf{12.3} & 45.1 & \textbf{16.8} & \textbf{27.2} & \textbf{48.2} & \textbf{23.4} & \textbf{28.4} & 42.1 & 13.5 && 35.1\\    
& WHDMM~\cite{RGBD_wang2016action} && \textbf{78.5} & \textbf{84.5} & \textbf{62.7} & \textbf{64.7} & \textbf{66.1} & 12.3 & 17.2 & \textbf{72.3} & 67.9 & 20.1 && 12.5 & 11.7 & \textbf{61.1} & 10.1 & 16.7 & 22.5 & 17.0 & 11.2 & \textbf{71.5} & \textbf{23.5} && \textbf{40.2}\\
\midrule
\multirow{3}{*}{SK-K} &TCN\cite{skeleton_tcn} && \textbf{91.2} & 82.0 & \textbf{71.4} & \textbf{86.0} & \textbf{92.2} & \textbf{91.7} & \textbf{87.6} & \textbf{93.0} & \textbf{89.2} & \textbf{92.6} && \textbf{57.5} & \textbf{76.0} & \textbf{92.9} & \textbf{87.8} & \textbf{66.8} & \textbf{70.5} & \textbf{95.0} & \textbf{76.1} & \textbf{76.1} & 76.4 && \textbf{82.6}\\   
& TSRNN~\cite{skeleton_tsrnn} && 90.0 & \textbf{85.0} & 70.6 & 81.0 & 91.0 & 90.5 & 86.6 & 91.8 & 86.6 & 91.4 && 56.7 & 75.1 & 91.7 & 86.8 & 66.0 & 69.7 & 93.8 & 75.1 & 65.1 & \textbf{85.4} && 81.5\\
& STGCN~\cite{skeleton_stgcn} && 90.6 & 83.5 & 71.0 & 83.5 & 91.6 & 91.1 & 87.1 & 92.4 & 88.7 & 92.0 && 57.1 & 75.6 & 92.3 & 87.3 & 66.4 & 70.1 & 94.4 & 75.6 & 75.6 & 75.9 && 82.1\\
\midrule
\multirow{3}{*}{SK-V} & TCN~\cite{skeleton_tcn} && 82.1 & \textbf{77.2} & \textbf{67.2} & \textbf{87.2} & \textbf{83.8} & 83.3 & 80.1 & 84.4 & \textbf{81.4} & 84.0 && 36.0 & 50.9 & \textbf{64.3} & 60.3 & 43.4 & 46.4 & 66.0 & 50.9 & 50.9 & 51.1 && 64.1\\
& TSRNN~\cite{skeleton_tsrnn} && \textbf{83.0} & 77.2 & 67.1 & 77.4 & 82.1 & \textbf{84.4} & \textbf{80.5} & \textbf{84.9} & 79.9 & \textbf{84.1} && \textbf{38.4} & \textbf{64.1} & 58.3 & \textbf{64.0} & \textbf{46.3} & \textbf{49.4} & \textbf{70.1} & \textbf{54.1} & \textbf{64.1} & \textbf{64.3} && \textbf{67.5}\\
&STGCN\cite{skeleton_stgcn} &&  57.7 & 53.2 & 45.2 & 53.2 & 58.4 & 58.0 & 55.5 & 58.9 & 56.5 & 59.6 && 36.4 & 48.2 & 58.7 & 55.6 & 42.3 & 44.6 & 60.1 & 45.2 & 25.2 & 54.3 && 50.7\\
\midrule
\multirow{1}{*}{EMG} & LSTM-FFT && 72.3 & 51.6 & 35.1 & 54.8 & 90.6 & 40.0 & 30.3 & 36.6 & 11.9 & 72.8 && 51.2 & 56.5 & 16.1 & 41.6 & 17.3 & 48.4 & 45.7 & 31.4 & 46.2 & 33.0 && 44.1\\
\midrule
\multirow{2}{*}{SK-K-E} & TCN-RMS && 91.1 & 83.0 & \textbf{73.4} & \textbf{88.0} & 93.2 & \textbf{94.7} & \textbf{87.8} & 91.0 & \textbf{91.4} & \textbf{95.6} && 60.5 & \textbf{79.8} & \textbf{91.9} & 88.8 & \textbf{70.8} & \textbf{72.5} & 94.0 & 74.1 & \textbf{78.1} & 74.4 && 83.6\\
& TCN-FFT && \textbf{92.0} & \textbf{83.7} & 72.1 & 85.7 & \textbf{94.0} & 93.5 & 87.3 & \textbf{94.8} & 91.0 & 94.4 && \textbf{60.6} & 78.5 & 91.3 & \textbf{89.6} & 70.1 & 71.9 & \textbf{94.8} & \textbf{79.5} & 77.6 & \textbf{77.9} && \textbf{84.0}\\
\midrule
\multirow{2}{*}{SK-V-E} & TCN-RMS && \textbf{86.7} & \textbf{80.7} & \textbf{70.3} & \textbf{87.9} & \textbf{87.1} & \textbf{84.5} & \textbf{83.6} & \textbf{85.1} & \textbf{82.1} & 83.6 && \textbf{63.5} & \textbf{51.6} & 64.4 & \textbf{60.3} & \textbf{45.4} & 46.0 & \textbf{65.8} & 50.5 & 51.2 & 51.1 && \textbf{69.1}\\
&TCN-FFT && 82.2 & 77.5 & 67.3 & 87.3 & 83.8 & 83.4 & 80.5 & 84.7 & 81.7 & \textbf{84.5} && 37.0 & 51.4 & \textbf{64.5} & 60.0 & 43.5 & \textbf{47.4} & 64.0 & \textbf{53.9} & \textbf{52.9} & \textbf{51.1} && 66.8\\
\bottomrule
\end{tabular}
\label{tbl:result}
\end{table*}

\subsection{Skeleton-Kinect}
We introduce the action recognition baselines based on skeleton modal in this section.

\noindent\textbf{Temporal Convolutional Networks (TCN)}~\cite{skeleton_tcn} learns an interpretable spatio-temporal representation. To train the model on the Kinect Skeleton modality, we modify the data as the same format of NTU-RGB-D~\cite{SK_data_UTN}. \textbf{Two Stream Recurrent Neural Network (TSRNN)}~\cite{skeleton_tsrnn} provides an RNN framework with two-stream structure to explore spatial configurations and temporal dynamics for action classification. We process the data in the same way as TCN. The batch size is set to $256$, the maximum iteration number is set to $2,000$, and the learning rate is set to $0.02$. \textbf{Spatial Temporal Graph Convolution Network (STGCN)}~\cite{skeleton_stgcn} learns both the spatial and temporal patterns from data simultaneously. It overcomes the limited expressive power and difficulties of generalization. The data is processed in the same way as TCN. We train the model with $80$ epochs, using SGD as the optimizer.

\subsection{Skeleton-Vicon}
Vicon system captures skeleton data with higher localization quality. In our evaluation, we deploy the same baselines as (Skeleton-K) while modify the data format to satisfy the requirements of each baselines. For \textbf{TCN}~\cite{skeleton_tcn} approach, we change the spatial connection graph from $25$ joints to $39$ joints. Vicon data contains higher frame rate, and we also increase frames for other models. The remaining parameters are kept the same. The dimension of the feature is increased to $273$. \textbf{TSRNN}~\cite{skeleton_tsrnn} needs the part of the body (\ie, one trunk, two legs, and two arms, as well as whole body). We use the index groups of different body parts via the $39$ joints of Vicon and keep other parameters be consistent. \textbf{STGCN}~\cite{skeleton_stgcn} needs a joint adjacency graph. Thus, we generate the connection graph for $39$-joint while other parameters are consistent.

\subsection{Skeleton-(Kinect/Vicon) \& EMG}
To prove the effectiveness and complementariness of EMG modal, we combine EMG with skeleton modalities together in low-level domain.

\noindent\textbf{TCN-RMS.} We first obtain the EMG features. The time window has the same size as the sampling time between two frames of the Skeleton-K. We then concatenate the RMS directly with the skeleton data and input the combination data to TCN~\cite{skeleton_tcn} for classification. The hyper parameters we used are the same as the parameters aforementioned. And we find out that such combination features have improved the performance on the action recognition task. \textbf{TCN-FFT.} FFT-based feature merging strategy is also evaluated since EMG is a temporal signal which can be explored in frequency space. Similar as TCN-RMS, we set a time window to extract the frequency distribution feature of the signal in each channel, and forward to classifiers.

\subsection{Results and Analysis}
We used top-1 accuracy to evaluate each baseline (Table \ref{tbl:result}). The left column shows the modals utilized for evaluation. \textit{SK-K} and \textit{SK-V} indicate the single-view skeleton data Skeleton-K and Skeleton-V respectively. \textit{SK-K-E} and \textit{SK-V-E} denote the multi-modal setting where EMG modal is combined with Skeleton-K and Skeleton-V respectively. For RGB modality, TSN outperformed other benchmarks with an accuracy of $74.8\%$. For each action, we noticed that some actions were easy to recognize and received over $90\%$ accuracy (\ie, \textit{Box}, \textit{Raise Hand}, and \textit{Move Table}). However, \textit{Wave Hand} only got $19.8\%$ accuracy. This is because this kind of action has low visual distinctiveness especially when the subjects wear black suit in data collection procedure. For depth modality, WDMM obtained $35.1\%$ average accuracy, while WHDMM greatly outdid that with $40.2\%$. Compared with the $82.6\%$ accuracy obtained from the skeleton data alone, the added EMG signal improved this by $1.4\%$ (\ie, TCN-FFT with $84.0\%$). Thus, we conclude that the actions with slightly movements (\ie, \textit{Check Watch}, \textit{Answer Phone}) have been improved with EMG features. Since the EMG signal can react to the action without obvious motion, while the visual feature is indistinctive. As a consequence, the result is reasonable. The reason why EMG with FFT does not have significant improvement may be that the FFT features are complicated and significantly different compared with skeleton motion structure. If these two modalities are simply and directly concatenated together, the extra dynamic information of FFT features could not be fully utilized. The EMG with LSTM-FFT also proves EMG is useful. The accuracies of several actions are extremely high (\ie, \textit{Jump}). However, similar actions(\ie, \textit{Wave Hand} and \textit{Raise Hand}; \textit{Stand Up} and \textit{Sit down}), which can be easily classified by other modality, are sometimes hard for EMG. Therefore, a good fusion technique may help us get better results. We conjecture there are two reasons for the relatively low baseline performances of Vicon data. (1) Some missing points make the data more challenging. (2) The generation strategy of spatial graph for Vicon data is different from the Kinect skeleton model default setting. Comparing our skeleton dataset with NTU-RGBD~\cite{SK_data_UTN} dataset, which included the same skeleton baselines as ours, we consider the differences in scores are justifiable in two-fold. (1) Our dataset contains less but more challenging action clips. (2) We only utilized 3D reference frame from the skeleton modality, while \cite{SK_data_UTN} fully utilized more motion information such as orientation. Regardless, the evaluation of our dataset can be further boosted with the added non-visual modality, the EMG signal. We believe more advanced feature extraction methods and multi-modality fusion strategies could further improve the learning performance. To this end, a lot of open questions and challenges are left for future exploration.

\section{CONCLUSION}
We have introduced a new multi-modal human action dataset in this paper which is called EV-Action dataset. The proposed dataset consists of RGB, depth, skeleton, and EMG data. All modalities have been labeled and aligned across $7,000$ samples collected from $70$ human subjects. In general, EV-Action has two major  advantages over the other action-based video collections. ($1$) we have utilized an optical tracking based Vicon system to capture more accurate and comprehensive skeletal data; ($2$) we have introduced a non-visual EMG modality associated with other visual modalities. We also have provided several state-of-the-art benchmarks for each modality to prove the effectiveness. Moreover, we have designed effective and simple approach for EMG-based action recognition task and achieved highest performance. Further, the experiments  have demonstrated that the effective and complimentary information is extractable from EMG for human action analytical tasks. Overall, EV-Action can serve in widespread research and applications concerning human motion understanding. We hope EV-Action can have a significant impact on motion understanding, computer vision, biomechanics, and other interdisciplinary areas.

                                  
\newpage                                  

\bibliographystyle{ieee}
\bibliography{ref_FG}
\end{document}